\pdfoutput=1

\documentclass[11pt]{article}

\usepackage{ACL2023}

\usepackage{times}
\usepackage{latexsym}
\usepackage{algorithm}
\usepackage{algorithmicx}
\usepackage[noend]{algpseudocode}
\usepackage[T1]{fontenc}
\usepackage[utf8]{inputenc}
\usepackage{amsmath,amssymb}
\usepackage{mathtools}
\usepackage{graphicx}
\usepackage{booktabs}
\usepackage{paralist}
\usepackage{tabularray}
\usepackage{microtype}
\usepackage{inconsolata}
\usepackage{acronym}
\usepackage{placeins}
\usepackage{makecell}
\usepackage{etoolbox}

\title{A Modular Approach for Multimodal Summarization of TV Shows}

\author{Louis Mahon \\
  School of Informatics \\
  University of Edinburgh \\
  Edinburgh, UK \\
  \texttt{lmahon@ed.ac.uk} \\\And
  Mirella Lapata \\
  School of Informatics \\
  University of Edinburgh \\
  Edinburgh, UK \\
  \texttt{mlap@inf.ed.ac.uk}
  \\}

\begin{document}
\acrodef{mdl}[MDL]{minimum description length}
\acrodef{llm}[LLM]{large language model}

\maketitle
\begin{abstract}
In this paper we address the task of summarizing television shows, which touches key areas in AI research: complex reasoning, multiple modalities, and long narratives. We present a modular approach where separate components perform specialized sub-tasks which we argue affords greater flexibility compared to end-to-end methods.  Our modules involve detecting scene boundaries, reordering scenes so as to minimize the number of cuts between different events, converting visual information to text, summarizing the dialogue in each scene, and fusing the scene summaries into a final summary for the entire episode. We also present a new metric, \textsc{PRisma} (\textbf{P}recision and \textbf{R}ecall Evaluat\textbf{i}on of \textbf{S}ummary F\textbf{a}cts), to measure both precision and recall of generated summaries, which we decompose into atomic facts. Tested on the recently released SummScreen3D dataset \cite{papalampidi2023hierarchical3d}, our method produces higher quality summaries than comparison models, as measured with ROUGE and our new fact-based metric, and as assessed by human evaluators. Code for our experiments and metric are at \url{https://github.com/Lou1sM/modular_multimodal_summarization}.
\end{abstract}

\section{Introduction}

In this paper, we address the challenging task of summarizing
television shows which has practical utility in allowing viewers to
quickly recall plot points, characters, and events without the need to
re-watch entire episodes or seasons. From a computational standpoint,
the task serves as a testbed for complex reasoning over long
narratives, involving multiple modalities,
non-trivial temporal dependencies, inferences over events, and multi-party
dialogue with different styles. An added difficulty concerns assessing
the quality of generated summaries for long narratives, whether
evaluations are conducted by humans or via automatic metrics.

Most prior work on creative summarization does not consider the above
challenges all at once, focusing either on the text modality and
full-length narratives with complex semantics
\cite{gorinski-lapata-2015-movie,chen-etal-2022-summscreen,agarwal-etal-2022-creativesumm}
or on short video clips which last only a couple of minutes
\cite{tapaswi2016movieqa,lei-etal-2018-tvqa,Liu2020ViolinAL}. A
notable exception are \citet{papalampidi2023hierarchical3d}, who
incorporate multimodal information into a pre-trained textual
summarizer by adding (and tuning) adapter
layers~\cite{rebuffi2017learning,houlsby2019parameter}. On the
evaluation front, there is no single agreed-upon metric for measuring
summary quality automatically, although there is mounting evidence
that ROUGE \cite{lin-2004-rouge} does not discriminate between
different types of errors, in particular those relating to factuality
\cite{min-etal-2023-factscore,clark-etal-2023-seahorse}.

\begin{figure*}
    \centering
    \includegraphics[trim={1cm 0 1cm 0}, width=.87\textwidth]{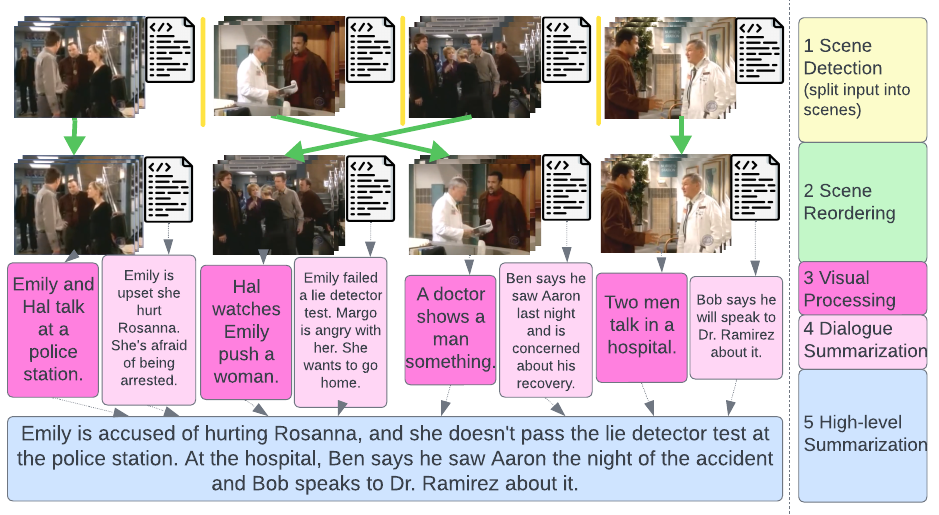}
    \vspace{-.2cm}
    \caption{Graphical depiction of our approach for long-form multimodal summarization where different subtasks are performed by five,
specialized modules (shown in different colors).  We use simplified summaries for display and show only four scenes. This full episode (\textit{As the World Turns} aired 01-06-05, contains 29 scenes.}
    \label{fig:method}
\end{figure*}

While end-to-end models are a popular choice for summarization tasks
\cite{chen-etal-2022-summscreen,zhang:ea:2020}, more modular approaches
have been gaining ground recently
\cite{guan-padmakumar-2023-extract,Gupta2022VisualPC,sun2023pearl} for
several reasons. Modules can be developed independently, and exchanged
for better versions if available, new modules can be added to create
new solutions or repurposed for different tasks, and dependencies
between modules can be rearranged. Aside from greater controllability,
modular approaches are by design more interpretable, since errors can
be inspected and attributed to specific components.  In this paper we
delegate the end-to-end task of summarizing from multiple modalities
(i.e.,~TV show video and its transcript) to more specialized modules,
each responsible for handling different subtasks and their
interactions. Our approach is depicted graphically in Figure~\ref{fig:method}.

As scene breaks are not always given explicitly, we devise an
algorithm to identify them from the order of the speaker
names (row~1, Figure~\ref{fig:method}). Additionally, we select the optimal order in which to
re-arrange scenes (row~2, Figure~\ref{fig:method}), as these often appear  non-linearly
(e.g.,~cuts between different subplots or flashbacks).  Next, we
produce summaries in a two-layer process. A vision-processing module \cite{lei-etal-2020-mart,lin2022swinbert}  converts the video to text using visual captioning (row~3, Figure~\ref{fig:method}), which leverages the strong specialized ability of vision-to-text models and allows us to treat the problem as one of text-to-text summarization. We also summarize each transcript
scene independently with a module specialized for dialogue
summarization (row~4, Figure~\ref{fig:method}). Finally, a high-level summarization module specialized
for narrative summarization fuses the sequence of scene summaries
into a final summary (row~5, Figure~\ref{fig:method}).


We also propose a new metric for assessing the factuality of generated
summaries by adapting FActScore \cite{min-etal-2023-factscore}, a recently introduced
metric for detecting hallucination in text
generation. We break the generated summary into atomic facts, and
check what fraction of them are supported by the reference. This we
term fact-precision. We also do the same in reverse, breaking the reference into facts and measuring what fraction are supported by the
generated summary, which we term fact-recall. Our metric, \textsc{PRisma}
(\textbf{P}recision and \textbf{R}ecall Evaluat\textbf{i}on of \textbf{S}ummary
F\textbf{a}cts), is the harmonic average of these two scores.
%
%
Our contributions are:
\begin{itemize}
  \itemsep0em 


\item We present a novel modular approach to multimodal summarization,
  where separate subtasks are performed by separate modules;

\item Our  modules involve detecting scene breaks, reordering
  each scene, summarizing the dialogue therein, converting the
  visual information to text, and fusing the scene-summaries into a
  final summary (see Figure~\ref{fig:method});

 \item We present two novel algorithms, for determining the optimal
      order in which to place each scene, and for identifying where
      the scene breaks are located in the transcript;

    \item We devise a new metric for summarization, based on splitting
      text into atomic facts, which captures both precision and
      recall and correlates significantly with human judgments.
\end{itemize}


\section{Related Work} \label{sec:related-work}
In the area of long-form summarization, various methods have been
proposed to deal with inputs that exceed the context size of a
\ac{llm}.  Memwalker \cite{chen2023walking} forms a tree of
hierarchical summary nodes and traverses it during inference to
find the most relevant parts of the input
text. \citet{pang-etal-2023-long} propose a two-layer method where
the top layer captures coarse long-range information and produces
top-down corrections to a local attention mechanism in the lower
layer. \citet{chang2023booookscore} describe two settings
for long-form summarization with \acp{llm}: hierarchical merging
 summarizes chunks of the input sequentially,
whereas iterative updating continually updates a single summary for
each of the chunks. We also adopt a two-layer approach but differ in that we divide the input into semantically meaningful chunks, i.e.,~scenes, rather than uniform chunks, and we summarize each independently before fusing them together. 


The
problem of generating descriptions for videos has also received
much attention, largely independently from long-document summarization.
It is common practice \cite{zhang2021open, pan2020spatio, ye2022hierarchical} to 
extract features from each frame individually, then fuse them into
a single feature vector
and decode with a
language model.  Swinbert \cite{lin2022swinbert} instead uses an
end-to-end video network, dispensing with image-based encoders, and
samples frames densely. \citet{lei-etal-2020-mart} generate 
descriptions for short videos with a memory-augmented vision transformer. Popular video captioning datasets  \cite{chen2011collecting,xu2016msr} are only
${\sim}10$s in length.  YouCook \cite{ZhXuCoCVPR18} is a recent dataset with longer videos (5min on
average), but  still far shorter
than the TV shows we focus on here.

Multimodal summarization is an extension of video captioning in which
the input contains text as well as video. \citet{pan2023kosmos} tackle
the analogous problem for still-images with a single model whose
architecture allows image and text input and can produce a text
description as output. \citet{bhattacharyya-etal-2023-video} convert
the video to text and feed it to a visual storytelling
model. \citet{tsimpoukelli2021multimodal} train a vision encoder to
produce a sequence of vectors which, when fed to an \ac{llm}, produce a
textual description of the image
contents. 
\citet{papalampidi2023hierarchical3d} apply a similar idea
to multimodal summarization, extracting a feature vector
from the visual input which is fed, along with the token embeddings
from the transcript, to a summarization network. Our method does not try to extract a visual feature vector that functions like a token embedding, but rather extracts actual tokens, i.e.,~a textual description, from the video.

\section{Decomposition into Modules}
\label{sec:decomp-mult-summ}

Our decomposition of the summarization task into modules is motivated
by three assumptions specific to TV shows: (1) each scene is somewhat
self-contained and, on a coarse level, the events it depicts can be
understood independently of other scenes (2) the order in which scenes
appear is not necessarily the optimal order to facilitate
understanding, often shows cut back and forth between different
plotlines, and sometimes they are presented non-linearly (3) an
effective way to capture visual information in a multimodal text
summary is to translate it into natural language.  These
assumptions motivate our modular approach, which is depicted
graphically in Figure~\ref{fig:method}. Five separate subtasks are
performed by separate components: scene-break detection (top row),
scene reordering (second row),
converting visual information to text (third row), dialogue summarization (fourth row), and high-level
summarization (fifth row).

Assumption (1) motivates our choice to detect scene breaks and
summarize each independently with a module specialized for dialogue
summarization and then later fuse these with a high-level
summarization module to produce a final output summary. Assumption (2)
motivates our scene-reordering algorithm: we do not simply concatenate
scene-level summaries in the order in which they appear, but rather we
compute an optimal order designed to minimize the number of
transitions between different plotlines. Assumption (3) motivates our
choice of how to capture the visual information in our summaries.  A visual processing module produces a textual description of
the video for each scene, which is fed, alongside the dialogue
summaries to the high-level summarization module. As a result,
 the high-level and  dialogue
summarization modules only need to focus on the single modality of
text. 


\subsection{The Multimodal Summarization Task}
\label{sec:mult-summ-task}

Before discussing the details of the various modules, we provide
information on our specific task and the dataset we are working
with. We develop and evaluate our approach on SummScreen3D\footnote{\url{https://github.com/ppapalampidi/video_abstractive_summarization}}
\cite{papalampidi2023hierarchical3d}, which to our knowledge is the
only existing dataset for long-form video summarization. It consists
of 5,421 videos of TV episodes (mostly soap operas) varying in length
from \mbox{30--60min}, with accompanying transcripts (on average 6K tokens long) and summaries
that were written by fans and scraped from public websites (average length is~200 tokens). These are
partitioned into 296 each for validation and testing, and the
remaining 4,829 for training. Videos can have multiple summaries from
different fansites (the average number of summaries per episode is~1.53), giving a total of 8,880 training pairs. 

In SummScreen3D, each data point contains a video, a transcript that
includes character names and, sometimes, marked scene breaks, and a
closed captions file, which is used to display subtitles and has
timestamps but not speaker names. We align the lines in the transcript
with those from the closed captions. These do not match perfectly,
because of slight errors in the automatic transcription. For each line
$t$ in the transcript and utterance~$c$ in the caption, we estimate a
similarity score as $\tfrac{|f(t,c)|}{\min{(|t|,|c|)}}$, where $f$
computes the longest common subsequence. Then we use dynamic time warping
to align both sequences \cite{myers1981comparative,
  papalampidi2021movie}. As a result, we obtain alignments of
transcript utterances to video segments, which allows us to use the scene-breaks from the former to segment the latter. 

\subsection{Scene Detection} \label{subsec:scene-detection}
This module (Figure~\ref{fig:method}, row~1) partitions the transcript into contiguous chunks.  Each
line in a transcript begins with the name of the character speaking,
and our algorithm seeks a partition where each chunk contains only
a small number of characters.

We define a cost for a given partition, by invoking the
\ac{mdl} principle \cite{grunwald2007minimum}, in which the optimal
representation of a piece of data is that which contains the
fewest of bits. \ac{mdl} has been shown to have many nice theoretical properties \citep{grunwald2007minimum} and has been used in recent machine learning applications from interpreting hidden representations \cite{voita2020information} to measuring noise-free information content in images \citep{mahon2024minimum, mahon2024towards}. We use it here to formalize the desirable properties of a good segmentation. 

Thus, the cost of a partition is the number of
bits needed to specify it.  We assume the total set of $N$ character
names for the entire transcript is given. Then, for each scene, we
make a scene-specific codebook for the $n$ characters that appear
there, which assigns each character a code as an index from $0, \dots,
n-1$, which all have length $ \leq \lceil \log{n} \rceil$. Using
exactly $\lceil \log{n} \rceil$ guarantees a prefix-free
code \cite{grunwald1998minimum}\footnote{Because we do not need to convert to concrete codes, we
  omit the ceiling operator in our optimization objective.}. The number of possible codebooks with $n$ out
of $N$ characters is $N \choose n$. Imposing some order, e.g.,~lexicographic, on these, the number of
bits to specify one is
\[
C(N,n) \coloneqq \log{N \choose n} = \log{\frac{N!}{n!(N-n)!}}
\]
and the total cost of a given scene is then
\begin{equation} \label{eq:mdl-single-scene-cost}
C(N,n) + l \log{n}\,,
\end{equation}
where $N$ is the total number of speakers as before, and $n$ and $l$
are, respectively, the number of distinct speakers and the number of
transcript lines in the scene. See Appendix~\ref{app:scene-detection-details} for details and examples. Intuitively, the two terms in \eqref{eq:mdl-single-scene-cost} impose opposite pressures on the solution. Term 1, the bitcost of specifying a new code book, imposes a cost on each scene break, and means the algorithm only inserts a scene break if it leads to a sufficient reduction in term two to outweigh this cost. Term 2, the bitcost of using each code book to specify the characters in the corresponding scene, penalizes solutions where scenes contain lots of different characters, and encourages instead segmentations where the number of speaking characters in each scene is minimized. 

Let $m$ be the number of lines in the transcript. Then, for any $1\leq i < j \leq m$, the cost of all scenes from line $i$ to line $j$ under the optimal partition, written as $S(i,j)$, can be expressed recursively as 
\[
S(i,j) = \min_{2 \leq k \leq i} S(i,k) + S(k,j)\,,
\]
where $S(i,j)$ is defined as zero when $i=j$. This motivates a dynamic
programming solution, similar to the CYK algorithm for context-free
parsing \cite{kasami1966efficient}, in which we compute, in order, the
optimal solution of spans of lengths $2,\dots m$, and reuse solutions
of smaller spans when computing those of larger spans.
\begin{algorithm}[tb]
    \caption{Compute the optimal partition of transcript lines into
      scenes.} \label{alg:scene-split}
    \small
    \begin{algorithmic}
    \State $m \gets$ \# lines in the transcript
    \State $P \gets$ an $m \times m$ matrix, to store the cost of spans
    \State $Q \gets$ an $m \times m$ matrix, to store the optimal partition of spans
    \For{$i=2,\dots,m$}
        \For{$j=1,\dots,m-1$}
            \State $n \gets$ \# of unique characters in lines $j, \dots, j+i$
            \State $P[j,j+i] \gets C(N,n) +i\log{n}$
            \State $Q[j,j+i] \gets \emptyset$
            \For{$k=j,\dots,j+i$}
                \If {$P[j,k] + P[k,j+i] < P[j,j+i]$}
             \State $P[j,j+i] \gets P[j,k] + P[k,j+i]$
                    \State $Q[j,j+i] \gets Q[j,k] \cup \{k\} \cup Q[k,j+i]$
                \EndIf
            \EndFor
        \EndFor
    \EndFor
    \end{algorithmic}
\end{algorithm}
Our algorithm runs in $O(N^3)$, with only $O(N^2)$ calls to compute
the scene cost as per
Equation~\eqref{eq:mdl-single-scene-cost}. 
The scene breaks can be transferred to the video because of the dynamic time-warping alignment between the transcript and the timestamps in the closed captions, as described in Section~\ref{sec:mult-summ-task}.
Note, that our algorithm determines automatically the number of scenes. 

\subsection{Scene Reordering} \label{subsec:reordering}
We now discuss how we compute the optimal order of scenes  (Figure~\ref{fig:method}, row~2), prior to summarization.  We
first define a cost for a given order as
\begin{equation} \label{eq:order-cost}
    C(s_1, \dots, s_n) = \sum_{i=1}^{n-1} 1-\operatorname{IOU}(s_i, s_{i+1})\,,
\end{equation}
where $\operatorname{IOU}$ is the intersection over union of character
names. For example, if the characters in scene~$s_1$ are Alice and
Bob, and the characters in scene~$s_2$ are Bob and Charlie, then
\mbox{$\operatorname{IOU}(s_1,s_2) = \tfrac{1}{3}$}. Additionally, we
introduce a constraint that if the same character appears in two
different scenes, $s_i$ and $s_j$, then we should never swap the order
of $s_i$ and $s_j$, as that may violate causality between the two
scenes.

We approximately solve this optimization problem by
passing from $s_2$ to $s_n$ and moving each scene as far to the front
as possible without violating the causality constraint, if this leads
to an improved total order cost. We continue passing from $s_2$ to
$s_n$ until no changes are made.  Our algorithm runs in $O(N^2)$, where $N$ is
the number of scenes, typically ~30. The change in cost when moving a
scene depends only on the $\operatorname{IOU}$ cost of that scene and
its current and future neighbours, and the $\operatorname{IOU}$ cost
of all pairs of scenes can be cached.

\begin{algorithm}[tb]
    \caption{Minimize the number of transitions between scenes with different speakers.} \label{alg:scene-reorder}
\small
    \begin{algorithmic}
    \Function{IOU}{x,y}
        \State \Return $\frac{\text{\#characters in both x and y}}{\text{\#characters in x} + \text{\#characters in y}}$
    \EndFunction      
    \Function{Compute Optimal Order}{S}   
        \State $s_1, \dots, s_n \gets$ scenes in order of  appearance in~$S$; 
        \State $C \gets n \times n$ matrix; \Comment{cache for IOUs}
        \For{$i=1,\dots,n$}
            \For{j=i,\dots,n}
                \State $C[i,j] \gets 1 - \operatorname{IOU}(s_i,s_j)$;
                \If {$(s_i,s_j)==0$}
                    \State $C[j,i] = 1$
                \Else 
                    \State $C[j,i] = \infty$
                \EndIf
            \EndFor
        \EndFor
        \State $changed \gets True$; 
        \While {$changed$}
            \For{$i=2,\dots,2$}
                \State $n = \min\{j| C[j,i] < 1\} + 1$ \Comment{new idx for $i$}
                \State $cost1 \gets C[i-1, i+1] - C[i-1,i] - C[i,i+1]$;
                \State $cost2 \gets C[n-1, i] + C[i,n] - C[n-1,n]$;
                \If{$cost1 + cost2 < 0$}
                    \State move $s_i$ to position $new$ in $S$;
                    \State $changed \gets True$
                    \State break
                \EndIf
            \EndFor
        \EndWhile
    \EndFunction      
    \end{algorithmic}
\end{algorithm}


\subsection{Vision Processing}
\label{sec:vision-processing}

We explore two methods for the vision-processing module which differ
in architecture and scope (Figure~\ref{fig:method}, row~3).  SwinBERT \cite{lin2022swinbert} is a
dedicated video captioning model; it operates directly on a sequence of video frames for which it generates a
description. SwinBERT is end-to-end trained with a masked language
modeling objective, and a sparse attention mask regularizer for
improving long-range video sequence modeling. We apply SwinBERT to the video for each scene, sampled at 3 frames-per-second, to obtain a video caption, which we use as a description of video contents.

Kosmos-2 \cite{peng2023kosmos} is pretrained on several
multimodal corpora as well as grounded image-text pairs (spans from
the text are associated with image regions) and instruction-tuned on
various vision-language instruction datasets. Unlike SwinBERT, it operates on individual images, so we first extract three equally spaced \mbox{I-frames} from the h264 video encoding \cite{wiegand2003overview}, and take the captions from each.
We prompt Kosmos-2 with ``A scene from a TV show in which''.

Both SwinBERT and Kosmos-2 can generate uninformative textual
descriptions, e.g.,~``a man is talking to another man'' which we
discard. We also modify them to replace unnamed entities such as `the
man' with character names where these can be easily inferred. See
Appendix~\ref{app:cap-filtering} for details of our filtering and renaming procedures.

\subsection{Summary Generation}
\label{sec:summary-generation}

Because of our two-layer summarization approach, we can use a  relatively small backbone model, and are still able to encode very long input. In our experiments, we use variants of BART-large \cite{lewis2020bart} for both the dialogue summarization and high-level summarization modules, but any other similar model could be used instead.
For the dialogue summarization module (Figure~\ref{fig:method}, row~4), we use the public Huggingface checkpoint for BART which has been fine-tuned on the SamSum dataset \cite{gliwa-etal-2019-samsum}, to output multi-sentence summaries of dialogue. For the high-level summarization module (Figure~\ref{fig:method}, row~5), we
use BART, fine-tuned first for document summarization on the CNN/Daily Mail dataset \cite{DBLP:conf/nips/HermannKGEKSB15}, and then on SummScreen3D.
The input for the latter fine-tuning is our re-ordered scene summaries and visual
text descriptions, i.e.,~the output of the dialogue summarization,
vision-processing and reordering modules. The output is the gold
summaries from the SummScreen3D 
training set.

\section{Fact Precision, Fact Recall and \textsc{PRisma}}
\label{sec:fact-f1}

Hallucinations are a widely known issue with abstractive summarization
\cite{song-etal-2018-structure,maynez-etal-2020-faithfulness,kryscinski-etal-2020-evaluating,gabriel-etal-2021-go},
especially when the output is long \cite{min-etal-2023-factscore} and
information is being consolidated from multiple sources
\cite{lebanoff-etal-2019-analyzing}. Automated metrics  are crucial for our task and related
creative summarization applications where human evaluation is extremely
labor-intensive (e.g.,~watching long videos or reading
book-length transcripts), costly, and difficult to design
\cite{krishna-etal-2023-longeval}.

Our proposal is based on a recent metric, FActScore
\cite{min-etal-2023-factscore}, which aims to detect hallucination in
text generation. FActScore first parses
the generated text into atomic facts (i.e.,~short sentences conveying
one piece of information), and then determines whether these  are
\emph{supported} by an external knowledge source, such as
Wikipedia. The FActScore for some model output is the fraction of
the extracted facts that are judged to be supported. \citet{min-etal-2023-factscore}
recommend using InstructGPT for the first stage of converting the
text into facts, and Llama-7B \cite{touvron2023llama} for
checking whether these facts are supported (i.e.,~by zero-shot
prompting Llama to estimate whether a generated fact is
\texttt{True} or
\texttt{False}).\footnote{\url{https://github.com/shmsw25/FActScore}}

We adapt this for summary evaluation by replacing the external
knowledge source with the gold summaries. Note that this incorporates
relevance as well as accuracy. There are many true facts about the
TV show being summarized, and only some of them appear in the gold
summaries. We assume that facts that do not appear in the gold summary
are irrelevant and so should not be marked as correct if they appear
in our model summaries. This is our fact-precision metric. For
fact-recall, we do the same \emph{in reverse}: we convert the gold
summary into atomic facts, and then check whether each of these is
supported by the generated summary. Again, the score is the fraction
of such facts that are supported. A summary will get a high
fact-precision score if every atomic piece of information is both true
and relevant enough to appear in the gold summary. It will get a high
fact-recall score if it contains every atomic piece of information
that was contained in the gold summary. The final score for our
metric, which we term \textsc{PRisma} (\textbf{P}recision and \textbf{R}ecall
Evaluat\textbf{i}on of \textbf{S}ummary F\textbf{a}cts), is 
the harmonic mean of fact-precision $FP$ and
fact-recall $FR$:
\[
\operatorname{\textsc{PRisma}} = \frac{2}{\frac{1}{FP} + \frac{1}{FR}}\,.
\]

In our implementation, we used GPT4-Turbo for the extraction of atomic
facts \emph{and} the estimation of whether they are supported, as we
found that Llama-7B overestimates the number of supported facts.  We use the same prompts as 
\citet{min-etal-2023-factscore} and make a separate query, with
in-context learning examples, for each fact. In order to penalize
repetitive/redundant information, repeated facts are regarded as
unsupported. In the case that GPT indicates that the sentence is not
properly formed, we convert it to a single fact which is scored as
unsupported (see Appendix~\ref{app:prefs} for an example).

We examined whether \textsc{PRisma} correlates with human judgments of
factuality. Two annotators, both English native
speakers, were asked to watch four randomly selected shows from the
SummScreen3D development set. They were then shown facts generated by
GPT4 corresponding to summaries produced by several automatic systems
(those described in Section~\ref{sec:comparison-models}). For each
fact they were asked to decide whether it was supported by the episode
(1/\texttt{True} or 0/\texttt{False}). Human ratings significantly
correlate with GPT4's estimate on whether a fact is
supported (Pearson's \mbox{$r=0.5$}, \mbox{$N=520$}, \mbox{$p<0.01$}), with an inter-annotator agreement  of~$r=0.57$
(\mbox{$p<0.01$}) as upper bound.

\section{Experimental Evaluation} \label{sec:experimental-eval}

\begin{table*}
  \center
 \begin{tabular}{@{}ll@{~~}lr@{~~}ll@{~~}lccc@{}}
\toprule
 & \multicolumn{2}{c}{r1} & \multicolumn{2}{c}{r2} & \multicolumn{2}{c}{rlsum} & fact-prec & fact-rec & \textsc{PRisma}\\
\midrule
mistral-7b & 28.38 & (0.16) &  4.82 & (0.06) & 26.14 & (0.17) & 31.75 & 38.00 & 34.59 \\
llama-7b & 16.31 &  (3.31) &  1.99 & (0.75) & 14.40 & (2.96) & 16.10 & 26.60 & 18.84\\
central &40.42 & (0.27) &    9.04 & (0.29) &   38.13 & (0.27) & 34.57 & 31.41 & 32.91\\
startend & 32.53 & (0.38) &    6.37 & (0.20) &   31.80 &  (0.35) &  34.78 & 33.62 & 34.19\\
unlimiformer & 42.24 & (0.42) & 10.32 & (0.34) & 40.40 &  (0.42) & 37.31 &
\emph{47.69} & 41.87 \\ \midrule
adapter-e2e & 34.13 & (0.07) & 4.28 & (0.08) & 31.81 & (0.07)& 31.15 & 39.72 & 34.93 \\
modular-swinbert & \textbf{44.89} & \textbf{(0.64)} & \emph{11.39} & \emph{(0.22)} & \emph{42.92} & \emph{(0.59)} & \textbf{42.71} & 46.47 & \emph{44.51} \\
modular-kosmos & \emph{44.86} & \emph{(0.60)} & \textbf{11.83} & \textbf{(0.15)} &
\textbf{42.97} & \textbf{(0.56)} & \emph{42.29} & \textbf{48.54} &
\textbf{45.20}\\ \midrule
upper-bound &  \multicolumn{2}{c}{42.62}  & \multicolumn{2}{c}{9.87}  & \multicolumn{2}{c}{40.03} & 71.07 & 91.25 & 79.91\\
\bottomrule
\end{tabular}
\caption{Automatic evaluation results on SummScreen3D. The first block
  presents text-only models and the second one multimodal ones.  Best
  results are in \textbf{bold}, second-best in \emph{italics}. For
  ROUGE, we report the mean of five independent random seeds, with
  standard deviation in parentheses. We present example output in Appendix~\ref{app:example-summaries}.
  } \label{tab:main-results}
\end{table*}


Our full model was composed of the modules described in
Section~\ref{sec:decomp-mult-summ}. For those SummScreen3D transcripts without explicitly marked breaks (approximately 20\%), we split into scenes using Algorithm \ref{alg:scene-split}.
The high-level summarization module was trained for a max of 10 epochs, with early stopping using
ROUGE on the validation set, with a patience of 2. The optimizer was
AdamW with learning rate 1e-6, and a linear scheduler with 0 warmup
steps. Training and inference took place on a single NVIDIA A100-SXM-80GB GPU, taking seven and one hour(s), respectively. The other models in our framework are not fine-tuned.

\subsection{Comparison Models} \label{sec:comparison-models}
We compare our approach to the end-to-end model of
\citet{papalampidi2023hierarchical3d}, which
represents the state of the art on our task, and 
various text-only models operating on the transcript:
\paragraph{Unilimiformer} \cite{bertsch2023unlimiformer} is  a retrieval-based
method that is particularly suited to processing long inputs. It wraps around encoder-decoder transformers, with context size~$k$, and
can extend this size arbitrarily by storing an index of all input
tokens and replacing the cross-attention query mechanism with
the~$k$-nearest neighbours from this index. 

\paragraph{LLama-7B, Mistral-7B} As our backbone BART
models are relatively small, ${\sim}400$M parameters, we may be able
to obtain better summaries, simply using a larger model; we test this
hypothesis with LLama-7B and Mistral-7B, both fine-tuned for three
epochs on our training set.

\paragraph{startend} We implemented a simple baseline
which uses BART fine-tuned for dialogue summarization and takes scenes
from the start and end of the transcript up to the maximum that can
fit in the context size (1,024).  This is inspired by recent work
showing that even long-context transformers take information mostly
from the beginning and end of the input \cite{liu2023lost}.

\paragraph{central} is inspired by the method of
\citet{papalampidi2021movie}. Again it uses BART and selects a subset
of the input to fit in the context size. It computes a weighted graph
whose nodes are scenes and edge-weights are tf-idf similarity scores,
then uses the page rank algorithm to rank
scenes in order of importance, and selects the topmost important
scenes that can fit in the context window.


We also compare to an upper-bound which tests the gold
summaries against each other. Recall that SummScreen3D often contains multiple
summaries for each episode. We select one from the three websites with
the most uniformly-sized summaries and
treat it as if it was the predicted summary, then test it against the
remaining summaries.

\begin{table*}
  \center
\begin{tabular}{ll@{~~}ll@{~~}ll@{~~}cccc}
\toprule
 & \multicolumn{2}{c}{r1} & \multicolumn{2}{c}{r2} & \multicolumn{2}{c}{rlsum} & fact-prec & fact-rec & \textsc{PRisma}\\
\midrule
w/o transcript & 30.71 & (0.38) & 5.48 &  (0.15) & 28.64 &  (0.25) & 16.53 & 21.91 & 18.84\\
w/o video & 43.80 & (0.76) & 10.91 &  (0.15) & 41.67 & (0.72) & 40.45 & 41.47 & 40.95\\
w/o reordering & \textbf{44.98} & \textbf{(0.52)} & 11.61  & (0.13) & \textbf{42.97} & \textbf{(0.50)} & 39.82 & 45.17 & 42.33\\
w/o scene-detection & 44.57 &  (0.82) & 11.22 &  (0.34) & 42.56 & (0.83) & 37.91 & 39.52 & 36.70\\
full & 44.86 & (0.60) & \textbf{11.83} & \textbf{(0.15)} & \textbf{42.97} & \textbf{(0.56)} & \textbf{42.29} & \textbf{48.54} & \textbf{45.20}\\
\bottomrule
\end{tabular}
\caption{Effect of removing various modules from our method (modular-kosmos).} \label{tab:abl-results}
    \end{table*}

\subsection{Results}
\label{sec:main-results}

Table~\ref{tab:main-results} summarizes our results, as measured by ROUGE-1 (r1), ROUGE-2 (r2), ROUGE-Lsum (rlsum) (computed using the python-rouge package) and our new  \textsc{PRisma} metric. We present results with BERTScore \cite{zhang2019bertscore} in Appendix~\ref{app:additional-results}, for the sake of brevity. Results for
\citet{papalampidi2023hierarchical3d}, which we denote as  `adapter-e2e', were reproduced with their code. 
For \textsc{PRisma},  we use a
separate query to GPT for each fact to check whether it is
supported. There are roughly 70 facts per generated summary, which can
lead to significant financial cost if used excessively. Therefore,
while we report five random seeds for ROUGE, we select a single
seed (randomly) for fact-precision and fact-recall. 

\paragraph{Our modular approach outperforms comparison models across all metrics.} As shown in Table~\ref{tab:main-results}, our method, which we denote as `modular' significantly outperforms the previous end-to-end multimodal system of  \citet{papalampidi2023hierarchical3d}, and all comparison text-only models (upper block).  A two-sampled
t-test shows that improvement over the comparison models is
significant at $\alpha=0.01$ (calculation in Appendix~\ref{app:significance-calculations}).
We observe that billion-parameter models like Mistral and Llama struggle with this task (although Mistral is superior), despite being fined-tuned on SummScreen3D.  Amongst text-only models, Unlimiformer performs best which suggests that the ability to selectively process long context has a greater impact on the summarization task. As far as our model is concerned, we find that the type of visual processing module has an effect, allbeit small, on the quality of the output sumaries. Overall, \mbox{Kosmos-2} \cite{peng2023kosmos} has a slight advantage over SwinBERT \cite{lin2022swinbert} which we attribute to it having been trained on various image understanding tasks, besides captioning.



\paragraph{\textsc{PRisma} better reflects summary quality than ROUGE.}
Interestingly,  upper-bound ROUGE scores are
lower than those for several models. We regard this as a
shortcoming of ROUGE as a metric. Qualitatively, reading the different
gold summaries shows that they are more similar to and accurate with
respect to each other than any of the predicted summaries, including
ours. Although they differ in phrasing and length,
they all describe the same key events. 
\textsc{PRisma} better reflects this similarity and gives a much higher score
to `upper-bound' than to any of the models. The large gap to get
to the level of `gold upper-bound' reflects the difficulty of the
task. The comparison between ROUGE and \textsc{PRisma} suggests that the former
is useful for detecting which summaries are of very low quality, (e.g.,~if ROUGE-2 \mbox{is $<5$}). However, for distinguishing between multiple relatively decent
summaries, \textsc{PRisma} is more useful.

\paragraph{All modules are important, but the transcript is the most important of all.} Table \ref{tab:abl-results} shows the effect of removing four of the
five modules. In `w/o scene-detection', we remove the scene-detection
module and just split the input into equally-sized chunks equal to the
context size (1,024). In `w/o transcript' we do not include summaries
of the dialogue from the transcript, and the only input to the
high-level summarization module is the output of the visual processing
module. In `w/o video' we do the reverse: use only the dialogue
summaries without the output of the vision-processing module. In `w/o
reordering', we remove the scene-reordering module and present the
scene summaries to the high-level summarization module in the order in
which they appear. All these experiments are performed using the
Kosmos-2 captions (except `w/o video' which has no captions). 

Aside from ROUGE-1 and ROUGE-Lsum, which are roughly the same for `w/o reordering' and `w/o scene-detection', all ablation settings lead to a drop in all metrics. 
Unsurprisingly, the largest drop is in `w/o transcript' as most information is in the dialogue --- a TV show without sound or subtitles is difficult to follow. Interestingly, however, this setting still gets some n-grams (ROUGE) and facts (\textsc{PRisma}) correct, showing that our model is extracting useful information from the video. This is also clear from the `w/o video' setting, which shows a drop in all metrics. Qualitatively, we observe that 
the vision-processing module is most useful for identifying locations, e.g.,~`at the hospital'
which is generally not mentioned in the dialogue. For `w/o scene-detection' and `w/o reordering', the difference is moderate for ROUGE.
For 
\textsc{PRisma}, it is more substantial and  highly significant when taken over all facts in the dataset.

\paragraph{Our scene-detection algorithm is more accurate than uniform baselines.}
\label{subsec:scene-detection-accuracy}
Here, in Table \ref{tab:scene-split-acc}, we report an empirical test of the accuracy of our scene-detection algorithm (Section~\ref{subsec:scene-detection}, Algorithm~1). These figures are produced using an episode
which has explicitly marked scene breaks. We assign each transcript
line a label based on the scene it is in with respect to these
explicit scene breaks, e.g.,~all lines in the first scene get the label
`0'. Then we do the same for the scene breaks produced by our
algorithm, and compare the two sets of labels using unsupervised label
comparison metrics, as commonly used in clustering problems \cite{mahon2021selective}: accuracy
(ACC), normalized mutual information (NMI), and adjusted Rand index
(ARI), defined as, e.g.,~in
\citet{mahon2024hard}. We compare to two baselines. The first, denoted `uniform', divides each episode into $n$ equally-sized scenes, where $n$ is the average number of scenes per episode. The second, denoted `uniform oracle', does the same except sets $n$ to the true number of scenes in that episode.

Our method produces more accurate scene splits than both baselines, despite taking no information from the ground-truth labels. Moreover, many of the occasions on which it differs from the ground-truth scene splits appear at least as reasonable as the ground-truth splits when we observe only the character names in the transcript. This suggests that errors in the splits are not due to the algorithm itself but to the fact that, at present, it only uses character names, ignoring the speech itself. A future extension is to use named entities or all nouns from the speech as well as character names. 

\begin{table}[t]
\centering
    \begin{tabular}{lrrr}
\toprule
 & acc & nmi & ari \\
\midrule
ours & 0.890 & 0.881 & 0.766 \\
uniform oracle & 0.759 & 0.819 & 0.538 \\
uniform & 0.723 & 0.809 & 0.489 \\
\bottomrule
\end{tabular}
\caption{Accuracy of scene splits produced by our method, benchmarked against: `uniform' splits into average number of scenes in the dataset; `uniform oracle' into  ground-truth number of scenes for each episode.} \label{tab:scene-split-acc}
\end{table} 

\paragraph{Humans prefer summaries generated by our modular approach.} 
We selected 8 episodes from SummScreen3D, and asked human participants (recruited through the Prolific platform) to watch the episode, then rank the following summaries in order of perceived quality: ours, the gold summary, unlimiformer \cite{bertsch2023unlimiformer}, `adapter e2e` \cite{papalampidi2023hierarchical3d}, and finetuned Mistral.  We instructed participants to priortize informativeness and factual correctness, with a secondary emphasis on fluent well-written text. We included an attention-check question that participants would only be able to answer if they had watched the full episode, and asked for a short justification for the chosen ranking. The full instructions can be found in Appendix \ref{app:human-eval-guidelines}. 
We recruited five annotators for each of the eight episodes. 

Table \ref{tab:elo-scores} presents the ranking given to the systems by our participants. Specifically,  we calculate Elo (\citeyear{elo1967proposed}) scores for each of the summary methods we perform evaluation on. We adjust scores once after all comparisons are made so as to be independent of the order of episodes. We are interested only in relative scores and correlations with other metrics, which are independent of the adjustment factor $K$ and the starting score. So, we set $K=1$ and the starting score to $\tfrac{1}{2}$, and the score for each method reduces to the fraction of pairwise comparisons they won. We can also then observe that our \textsc{PRisma} metric is a much better predictor of human preference (Pearson's  $r = 0.792$) than ROUGE-2 ($r=0.594$) and ROUGE-Lsum ($r=0.512$). 

\begin{table}[t]
    \centering
\begin{tabular}{lc}
\toprule
Models & elo \\ \midrule
upper-bound &0.750 \\
modular-kosmos &  0.681\\
mistral-7b &  0.388\\
adapter-e2e & 0.353\\
unlimiformer & 0.319 \\
\bottomrule
\end{tabular}
  \caption{Ranking given to summarization systems by human subjects; Elo scores from pairwise comparisons.}
    \label{tab:elo-scores}
\end{table}

\section{Conclusion}
\label{sec:conclusion}

We addressed the task of summarizing TV shows from videos and dialogue transcripts. We proposed a modular approach where different specialized components perform separate sub-tasks. A scene-detection module splits the TV show into scenes, and a scene-reordering module places these scenes in an optimal order for summarization. 
A dialogue summarization module condenses the dialogue for each scene and a visual processing module produces a textual description of the video contents. Finally, a high-level summarization module fuses these into an output summary for the entire episode. We also introduced \textsc{PRisma}, a new metric for long-form summarization, based on splitting predicted and reference summaries into atomic facts. It captures both precision and recall, and correlates significantly with human judgments. 
In the future, we plan to test our method on even longer inputs, and explore settings where transcripts are not available.

\section*{Limitations}
While the modular approach we propose has advantages, such as allowing specialization of individual modules, and the ability to replace one module without affecting the others, it also has the disadvantage that it is difficult to fine-tune all modules. We fine-tune only the high-level summarization module, whereas for a monolithic end-to-end model, all parameters can be trained. 

Our proposed \textsc{PRisma} metric requires multiple calls to GPT which incurs a financial cost. We estimate that all the results presented in this paper cost about \$1300. This is still many times cheaper than hiring human evaluators to extract and score facts manually, which we estimate would take 400 person-hours and cost about \$15,000.

There is still a significant gap, in terms of \textsc{PRisma}, between our summaries and the upper bound of comparing the gold summaries to each other. This shows that the task is challenging and requires further advances to reach human-level. 



\section*{Acknowledgments}
We thank the anonymous reviewers for their feedback. We gratefully acknowledge the support of
Amazon and the UK Engineering and Physical Sciences Research Council (grant EP/W002876/1), and the feedback and advice from Liz Norred, Zhu Liu, Vimal Bhat and Rohith Mysore Vijaya Kumar.

\bibliography{bibliography}
\bibliographystyle{acl_natbib}

\appendix
\newpage 

\onecolumn 
\section{Example Summaries} \label{app:example-summaries}

Table \ref{tab:gold-bb-summ} shows the gold summary for an episode from the SummScreen3D testset. Tables \ref{tab:our-bb-summ} and \ref{tab:unl-bb-summ}--\ref{tab:nelly-bb-summ} show the summary of our model, and those of comparison models, respectively, for the same episode.

\FloatBarrier

\begin{table}[hbt]
\center
\small
\begin{tabular}{p{15cm}}
\begin{large}
   The gold Summary of \textit{The Bold and the Beautiful} episode (aired 05-05-06)
\end{large} \\
\toprule
\texttt{Ridge continues to beg Brooke to reconsider her decision to leave Forrester as Stephanie continues to voice her opinion. At Marone, Taylor pays Nick a visit. Nick is still angry about what Taylor implied when she disclosed that Brooke and Ridge slept together. Taylor tries to apologize and asks if things are all right between Nick and Brooke. Nick tells her that everything is fine and Brooke is quitting her job at Forrester. Taylor is unconvinced that Brooke will be able to let go of either Forrester or Ridge ! Brooke tells Ridge that she cannot fight with Stephanie any longer and that her future is with Nick. After kissing Ridge and saying that a part of her will always love him, she takes her things and leaves. Bridget and Dante are at home discussing Stephanie's interference in the custody of Dino. Bridget suggests that perhaps Eric can help them. Dante worries about what losing his job would do to his work visa. Bridget convinces him that because they all love Dino, they should be able to work something out. After some wine, Bridget reveals that she is ready to make love with Dante. As the two were in bed, Dante stops and reaches in the bedside drawer and presents Bridget with an engagement ring and pops the question. Brooke goes to see Nick at his office and tells him that she has left Forrester. Nick is pleased, although Brooke confesses that she hurt Ridge badly by walking out. Nick whisks her off to the Marone jet for a surprise getaway! At Forrester, Ridge angrily accuses Stephanie of causing all his problems with Brooke. Stephanie is stunned as Ridge bashes her with a vengeance and then clutching his chest, collapses to the floor!
}\\ \bottomrule
\end{tabular}
\caption{Gold summary from the SummScreen3D dataset for  \textit{The Bold and the Beautiful} episode (aired 05-05-06). Several summaries exist for this episode from different fansites. This summary is from TVMegaSite.} \label{tab:gold-bb-summ}
\end{table}

\begin{table}[hbt]
\center
\small
\begin{tabular}{p{15cm}}
\begin{large}
    Modular-Kosmos Summary (ours) of \textit{The Bold and the Beautiful} episode (aired 05-05-06)
\end{large} \\
\toprule
\texttt{Bridget and Dante plan to get married in Italy, but Bridget wants to spend more time with Dante. Stephanie tells Bridget that she wants to fire Dante and send him to Italy. Ridge tells Stephanie that he wants Brooke out of Forrester Creations. Brooke tells Nick that she is through fighting, she is moving to Paris. Nick tells her that she has to move out of the office. Nick says that he will not force her to work in Paris, but she says that she needs to move in. Ridge says that Nick is her future, and she has no choice but to leave. Stephanie says that Ridge needs her, but he does not want to accept her. He says that they are better off without her. She says that is not what she wants. She tells him that she does not know how he feels. She does not understand how he could do this to Brooke. He tells her to move on with her life. Taylor apologizes to Nick for suggesting that Ridge and Brooke slept together. Nick tells Taylor that he doesn’t let it affect his relationship with Brooke. Brooke tells Nick that she needs something to cheer her up, and Nick tells her that they are on their way to a tropical island. Ridge tries to convince Brooke to stay at Forrester Creations, but she tells him that she is leaving anyway. Taylor tells Ridge that she does not want to believe that he and Brooke have feelings for each other, but Ridge tells Taylor not to give up on her dreams. Nick and Brooke plan to take off on their trip, but Nick tells Brooke that he has a surprise for her, and that they will be going to a beautiful beach and a fruity drink. Brooke is very happy about Nick’s trip plans, but when she asks where they are going, Nick reveals that he wants to take her to a romantic beach.
}\\ \bottomrule
\end{tabular}
\caption{The summary generated by our model for \textit{The Bold and the Beautiful} episode (aired 05-05-06).} \label{tab:our-bb-summ}
\end{table}

\clearpage
\begin{table}[hbt]
\center
\small
\begin{tabular}{p{15cm}}

\begin{large}
    Unlimiformer Summary of \textit{The Bold and the Beautiful} episode (aired 05-05-06)
\end{large} \\
\toprule
\texttt{Brooke tells Ridge that Stephanie will never let her and Nick run Forrester Creations. She tells him that she is sick of her mother’s obsession with getting rid of her and she has had enough. Ridge tells his mother that she has been trying to get rid of Brooke for so long that it has become a compulsion for her. Bridget tells Dante that Stephanie cannot get away with this. She will call her dad and see what he will do. Dante tells Bridget that all she has to do is fire him and he will lose his work visa and go back to Italy. He needs a lawyer and Bridget says she will make sure he gets the best one. Dante thanks her for her support. She says he has no idea how much her support means to him.. Later, Dante opens a red velvet’.. Dante proposes to Bridget.. Source: tv.com tv.co.uk/soulmatestories: http://www.starz.com/dailymail.co/news/features/2013/01/25/24/ro-lorry-logan-says-that-if-he- walks-out-of-this-company-now they-may-never-get-back-what-they-had.html\#storylink\=cpy. Brooke tells him she still cares about him. He says he doesn’t care what his mother.  ..Source: TV.co  Back to Mail Online. Back to the page you came from: http: www. tv.cnn.com.com/. Back to the episode airs at 8 p.m. ET on Thursday, September 14, 2013.  The episode airs on ABC.  Click here to watch the full episode of The Young and the Restless.
}\\ \bottomrule
\end{tabular}
\caption{The summary generated by Unlimiformer \cite{bertsch2023unlimiformer} for \textit{The Bold and the Beautiful} episode (aired 05-05-06).} \label{tab:unl-bb-summ}
\end{table}

\begin{table}[hbt]
\center
\small
\begin{tabular}{p{15cm}}
\begin{large}
    Mistral-7B Summary of \textit{The Bold and the Beautiful} episode (aired 05-05-06)
\end{large} \\
\toprule
\texttt{Ridge and Brooke are having a conversation about their relationship. Ridge tells Brooke that he doesn't care what his mother thinks or wants anymore. He only cares about her. Brooke tells Ridge that Nick doesn't want her working at Forrester Creations. Ridge tells Brooke that Nick is telling her what to do now. Brooke tells Ridge that she is through fighting and that Stephanie has won. Dante and Bridget are having a conversation about Dante's work visa. Dante tells Bridget that he needs a lawyer. Bridget tells Dante that she will make sure he gets the best possible one. Dante tells Bridget that he wants to thank her for her support. Bridget tells Dante that he deserves to play a huge part in Dino's life. Bridget tells Dante that she will do whatever she can to make sure that happens. Nick and Taylor are having a conversation about Ridge and Brooke. Nick tells Taylor that he wants the containers off the ship. Nick tells Taylor that his people have schedules to keep. Nick tells Taylor that time and money are important. Taylor tells Nick that she wants to apologize. Taylor tells Nick that she still thinks she did the right thing telling him that Ridge and Brooke slept together. Taylor tells Nick that she never should have implied that Brooke wasn't being completely honest. Nick tells Taylor that he never believed her anyway.
}\\ \bottomrule
\end{tabular}
\caption{The summary generated by Mistral-7B for \textit{The Bold and the Beautiful} episode (aired 05-05-06).} \label{tab:mistral-bb-summ}
\end{table}

\begin{table}[!hbp]
\center
\small
\begin{tabular}{p{15cm}}
\begin{large}
    Adapter e2e Summary of \textit{The Bold and the Beautiful}  episode (aired 05-05-06)
\end{large} \\
\toprule
\texttt{Ridge: "You deserve to play a huge part in his life. And I'm going to do whatever I can to make sure that happens" "This is where we began. This is where our love affair began, Ridge. And it only seems fitting that this is where it should end" "I will never forget our accomplishments here. And part of me will always wonder what more we could have done together" "We'll leave L.A. and move to Paris. We'll work out of the office there -- the international office" "You wouldn't even be with him if my mother hadn't broken us up. Maybe we would still be married. We'd be raising our children." "I'm sorry, I know that you want me to be relieved. But on the other, I just don't think that -- I want you to be happy I've got a surprise for you" "It's really healed me, I guess. Because I wasn't ready to be with you." "You're the only one that's heartbroken if you got deported. After all, Dino wouldn't be the onlyOne that's Heartbroken if you got deported" "There's nothing anyone can say or do to change that now. You understand me? Good, I was hoping you'd say that. I never believed you anyway" "What do you want from me?" "I don't know what to say. I just want to be friends with you. I want to talk to you about everything." "What's your plan?" "We're going to Paris."
} \\ \bottomrule
\end{tabular}
\caption{The summary generated by the method of \cite{papalampidi2023hierarchical3d} for \textit{The Bold and the Beautiful} episode (aired 05-05-06).} \label{tab:nelly-bb-summ}
\end{table}

\twocolumn
\section{Caption Filtering and Inference of Names} \label{app:cap-filtering}
The output from the visual processing module for a given scene is a textual description of the visual information for that scene. Sometimes, this description is vague and merely adds noise to the high-level summarization module input. This is also clear from manually viewing the TV show episodes: many scenes are just headshots of characters talking, and do not convey plot information. For both SwinBERT and Kosmos-2, we filter out any captions that contain the following phrases
'a commercial', `talking', `is shown', `sitting on a chair', `sitting on a couch',  `sitting in a chair',  `walking around'. 
Addtionally, we replace occurrences of the phrase ``is/are seen'' with ``is/are''. This is because the captioning datasets these models are trained on often use the phrase ``is/are seen'', e.g.,~``a person is seen riding a bicycle'' instead of ``a person is riding a bicycle'', but we do not want such passive voice constructions in our summaries. 

\begin{figure}
    \centering
    \includegraphics[width=\columnwidth]{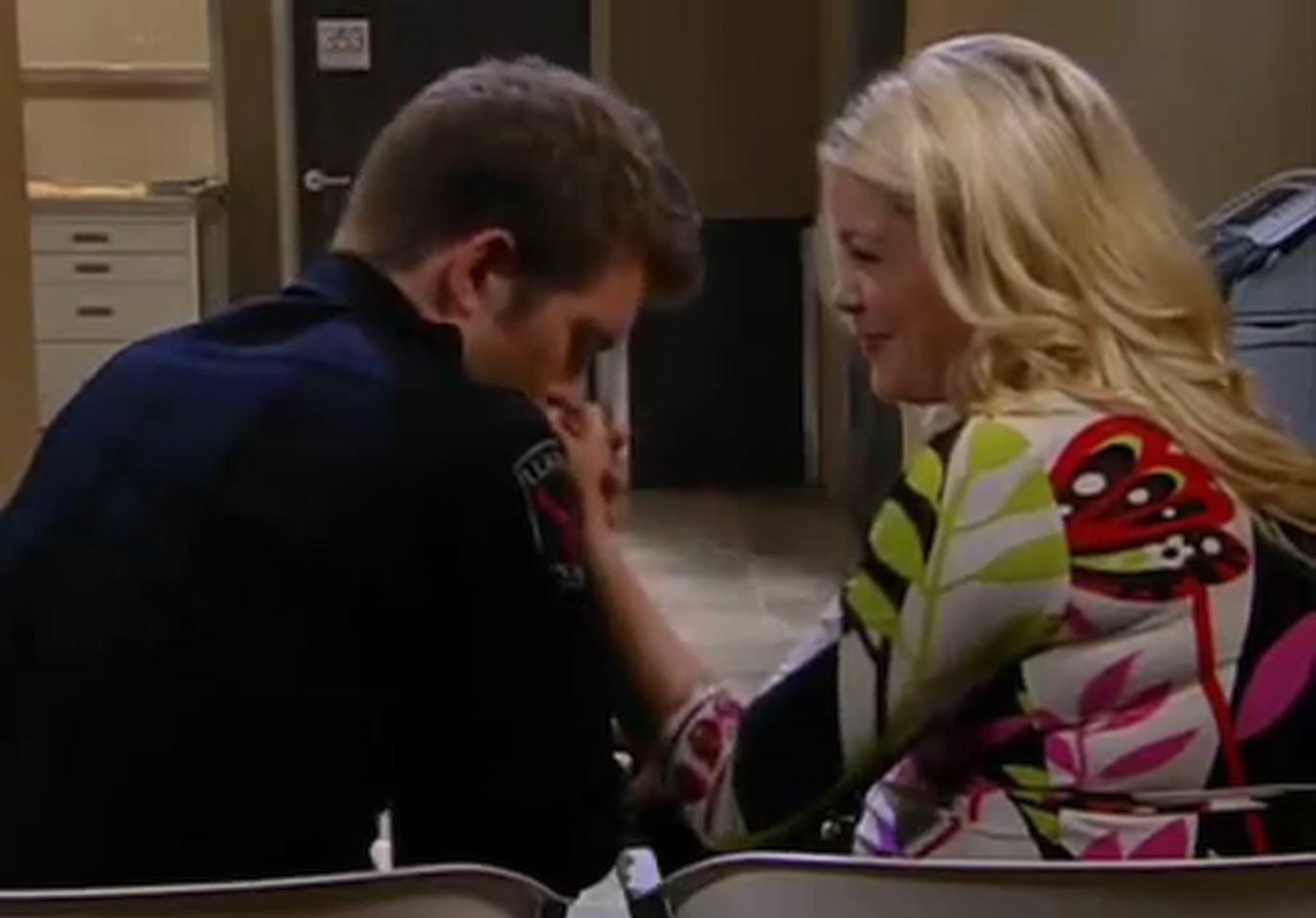}
    \caption{A selected keyframe from Scene 2 in \textit{One Life to Live}, (aired 10-18-10), which Kosmos-2 captions as ``a man is kissing a woman''. Our post-processing method to insert character names transforms this caption to ``Brody is kissing Jessica''.}
    \label{fig:brody-kissing-jessica}
\end{figure}

The captions do not contain character names, as these names cannot be inferred from the video alone. Therefore, for each sentence in the output of the visual processing module, we employ the following method to insert names, where they can be easily inferred. We categorize each name appearing in the transcript for the scene as male, female or neutral, using the vocabulary list for English names from Python's NLTK. Similarly, we assume noun-phrases ``he'', ``a man'', or ``a boy'' are male and  noun-phrases ``she'', ``a woman'', ``a girl'' are female. If there is only one male name, then we replace all male noun phrases with that name, similarly for female names. For example, from the caption for Scene 2 in \textit{One Life to Live}, (aired 10-18-10), as shown in Figure \ref{fig:brody-kissing-jessica}, the output of the visual processing module contains the sentence ``a man is kissing a woman'', and the only names in the transcript for that scene are ``Brody'', which is listed as male, and ``Jessica'', which is listed as female. Our method to insert names then transforms this caption to ``Brody is kissing Jessica''.

\section{\textsc{PRisma} Metric} \label{app:prefs}
Often, facts extracted by GPT4 are too vague and uninformative to be useful for assessing summary quality. Therefore, we perform a simple filtering to remove such facts. Specifically, we eliminate out any facts containing the following words for phrases: ``someone'', ``somebody'', ``something'', ``is a person'', ``are people'', ``is a character'', ``are characters''. Additionally, we eliminate all facts containing only two words, as we observe they are almost always uninformative and often are an ungrammatical use of a transitive verb in an intransitive context. Table \ref{tab:prefs-example} shows the full list of facts extracted from our summary for \textit{The Bold and the Beautiful}, (aired 05-05-06), showing which are eliminated, which are judged supported and which are judged unsupported.

\begin{table*}[t]
\center
\small
\begin{tabular}{p{15cm}} \toprule
\texttt{\textcolor{blue}{Bridget and Dante are planning to get married.} \textcolor{blue}{Bridget wants to spend more time with Dante.} \textcolor{red}{They plan to get married in Italy.} \textcolor{blue}{Stephanie wants to fire Dante.} \textcolor{blue}{Stephanie wants to send Dante to Italy.} \textcolor{red}{Stephanie tells Bridget.} \textcolor{red}{Bridget receives the information from Stephanie.} \textcolor{red}{Ridge wants Brooke out of Forrester Creations.} \textcolor{blue}{Ridge told Stephanie this.} Brooke tells Nick something. \textcolor{blue}{Brooke tells Nick that she is through fighting.} \textcolor{red}{Brooke tells Nick that she is moving to Paris.} Nick told her something. \textcolor{red}{Nick told her she has to move.} \textcolor{red}{She has to move.} \textcolor{blue}{She has to move out of the office.} \textcolor{red}{Nick told her she has to move out of the office.} \textcolor{blue}{Nick will not force her.} \textcolor{blue}{She will not work in Paris.} \textcolor{red}{She needs to move in.} \textcolor{red}{She needs to move to Paris.} \textcolor{red}{She will not work in Paris but she needs to move in.} \textcolor{red}{Ridge believes that Nick is her future.} \textcolor{red}{Ridge has no choice.} \textcolor{red}{Ridge must leave.} \textcolor{blue}{Stephanie believes Ridge needs her.} \textcolor{blue}{Ridge does not want to accept Stephanie.} \textcolor{blue}{Stephanie and Ridge have a relationship.} \textcolor{blue}{Stephanie and Ridge have a troubled relationship.} He says something. \textcolor{red}{He says that.} \textcolor{red}{He says that they are better off without her.} She says. \textcolor{red}{That is not what she wants.} She tells him something. \textcolor{red}{She tells him that she does not know.} \textcolor{red}{She tells him that she does not know how he feels.} \textcolor{red}{She does not know how he feels.} \textcolor{red}{She does not understand.} \textcolor{red}{He could do this to Brooke.} He tells her something. \textcolor{red}{He tells her to move on.} \textcolor{red}{He tells her to move on with her life.} \textcolor{red}{He does not have children.} \textcolor{red}{He is a proud uncle.} \textcolor{red}{He has four nieces and nephews.} Taylor apologized. \textcolor{blue}{Taylor apologized to Nick.} \textcolor{blue}{Taylor suggested that Ridge and Brooke slept together.} \textcolor{blue}{Ridge and Brooke are two people.} \textcolor{blue}{Ridge and Brooke slept together.} Nick tells Taylor something. Nick tells Taylor he doesn't let something affect his relationship with Brooke. \textcolor{blue}{Nick has a relationship with Brooke.} Nick's relationship with Brooke is affected by something. Nick doesn't let something affect his relationship with Brooke. Brooke tells Nick that she needs something to cheer her up. Nick tells her something. \textcolor{red}{Nick tells her they are on their way.} \textcolor{red}{They are on their way to a tropical island.} \textcolor{blue}{Ridge tries to convince Brooke.} \textcolor{blue}{Brooke is leaving.} \textcolor{blue}{Brooke is leaving Forrester Creations.} \textcolor{blue}{Brooke is leaving Forrester Creations anyway.} Taylor tells Ridge something. Taylor does not want to believe something. \textcolor{blue}{Ridge and Brooke have feelings for each other.} Ridge tells Taylor something. \textcolor{blue}{Taylor should not give up on her dreams.} \textcolor{blue}{Nick and Brooke plan to take off on a trip.} \textcolor{blue}{Nick has a surprise for Brooke.} \textcolor{red}{They will be going to a beautiful beach.} \textcolor{red}{They will be going to a beach.} \textcolor{red}{They will be going to a fruity drink.} \textcolor{red}{They will be going to a beautiful beach and a fruity drink.} \textcolor{blue}{Nick tells Brooke about the surprise.} \textcolor{blue}{Nick tells Brooke that they will be going to a beautiful beach.} \textcolor{red}{Nick tells Brooke that they will be going to a fruity drink.} \textcolor{blue}{Brooke is happy about Nick's trip plans.} \textcolor{blue}{Nick has trip plans.} \textcolor{blue}{Nick wants to take Brooke somewhere.} \textcolor{blue}{Nick wants to take Brooke to a beach.} \textcolor{blue}{The beach is romantic.} 
} \\ \bottomrule
\end{tabular}

\caption{\label{facts} Facts extracted with GPT4 from our summary of \textit{The Bold and the Beautiful}, (aired 05-05-06), from SummScreen3D. Facts that we filter out are written in black, those that are judged as supported by the gold summary are in \textcolor{blue}{blue}, while those unsupported are in \textcolor{red}{red}. After applying our filtering procedure, the number of facts reduces from \textbf{83} to \textbf{67}, of which 33 are judged supported and 34 unsupported, giving a factscore-precision of 49.25.} \label{tab:prefs-example}
\end{table*}

\section{Scene Detection Details and Examples} \label{app:scene-detection-details}

Here, we provide further details on the scene detection algorithm from Section \ref{subsec:scene-detection} and report a measure of its accuracy.

We calculate the cost for a given partition by assuming that the scene breaks under this partition and the placeholders where the speakers would go are given, and then asking how many bits are needed to fill in the speaker names. Then the receiver can infer the length of the scene-specific codebook as $2^{m}$, where $m$ is the length of each code in the scene. Note, we still need prefix-freeness, as just knowing the number of speaker lines in a scene doesn't allow us to distinguish between $10$ followed by $11$ and $101$ followed by $1$.

For example, consider the following sequence of character names from \textit{As the World Turns} (aired 01-09-07):
'Elwood', `Casey', `Elwood', `Casey', `Elwood', `Casey', `Elwood', `Casey', `Elwood', `Meg', `Paul', `Meg', `Paul', `Meg', `Paul', `Luke', `Meg', `Paul', `Luke', `Meg', `Luke', `Meg', `Luke', `Meg', `Luke', `Meg', `Paul', `Meg', `Luke', `Meg', `Luke', `Meg', `Luke', `Meg', `Luke', `Meg', `Luke', `Adam', `Gwen', `Adam', `Gwen', `Adam', `Gwen', `Adam', `Gwen', `Adam'. 

Abbreviating the character names as their first letters, the correct partition is 
\begin{enumerate}
    \item ECECECECE, 
    \item MPMPMPLMPLMLMLMLMPMPMLM- LMLML, 
    \item AGAGAGAGA
\end{enumerate}

Now consider the cost, as defined in Section \ref{subsec:scene-detection} of this partition. The total vocabulary size is 
\[
N = |\{E,C,M,P,L,A,G\}| = 7\,.
\]
For the first scene, $n_1=2$, $n_2=3$, $n_3=2$, $l_1=9$, $l_2=28$, $l_3=7$. Letting $c_i$ denote the cost for the $i$th scene, we then have
similarly,
\begin{align*}
    c_1 =& \log{7 \choose 2} + 9\log{2} \approx 13.392\\
    c_2 =& \log{7 \choose 3} + 28\log{3} \approx 49.508 \\
    c_3 =& \log{7 \choose 2} + 9\log{2} \approx 13.392\,,
\end{align*}
giving a total cost of approximately $13.392+49.508+13.392=76.292$.

Now compare this to the cost of a different partition of this sequence, say into scenes of uniform length:

\begin{enumerate}
    \item ECECECECEMPMPMP,
    \item LMPLMLMLMLMPMPM,
    \item LMLMLMLAGAGAGAGA
\end{enumerate}
In that case
\begin{align*}
    c_1 =& \log{7 \choose 4} + 15\log{4} \approx 34.392\\
    c_2 =& \log{7 \choose 3} + 15\log{3} \approx 28.167 \\
    c_3 =& \log{7 \choose 4} + 16\log{4} \approx 36.392\,,
\end{align*}
giving a total cost of approximately $34.392+28.167+36.392=98.951$. Therefore, because our algorithm computes the global minimum to this cost function, it is guaranteed to choose the correct partition over this uniform one. In fact, it selects the correct partition.

\section{Additional Results}
\label{app:additional-results}

\subsection{BERTScore} \label{app:bertscore}
Tables~\ref{tab:bertscore-comp-models} and \ref{tab:bertscore-ablation-settings} report results on summary quality according to  BERTScore  \cite{zhang2019bertscore}. Tables~\ref{tab:bertscore-comp-models} shows BERTScore for our model and comparison models, the analogue of Table \ref{tab:main-results} in the main paper, and Table~\ref{tab:bertscore-ablation-settings} shows BERTScore for the ablation settings, the analogue of Table~\ref{tab:abl-results} in the main paper. 

We find this metric does not distinguish well between the different settings, and scores all models very similarly. Even the ``w/o transcript'' setting, which qualitatively misses most of the important information in the episode, and scores poorly on ROUGE and \textsc{PRisma}, gets a high BERTScore. Moreover, even within each setting, the scores are very similar across different episodes, again in contrast to ROUGE, \textsc{PRisma} and qualitative evaluation (some episodes appear much easier to summarize than others). The average standard deviation across episodes, within each setting, is $2.2$, $1.39$, and $1.41$ for BERTScore-precision, BERTScore-recall and BERTScore-f1 respectively. In contrast, the same standard deviation for ROUGE-1, ROUGE-2, ROUGE-Lsum are $6.66$, $3.32$, $6.52$, while for factscore-precision, factscore-recall and \textsc{PRisma}, they are $14.92$, $15.14$ $14.94$, respectively.

This suggests that BERTScore always returns a similar score, regardless of the input. This inability of BERTScore to adequately distinguish different settings was also reported by \citet{papalampidi2023hierarchical3d}.

\begin{table}[t]
\begin{center}
\resizebox{\columnwidth}{!}{
\begin{tabular}{@{}llll@{}} \toprule
 & bs-precision & bs-recall & bs-f1 \\
\midrule
llama-7b & 75.48 (1.22) & 78.99 (0.21) & 77.13 (0.74) \\
mistral-7b & 79.85 (0.04) & 81.15 (0.07) & 80.47 (0.05) \\
central & 79.33 (0.39) & 82.46 (0.19) & 80.82 (0.24) \\
startend & 80.09 (0.98) & 82.58 (0.30) & 81.29 (0.54) \\
unlimiformer & 82.69 (0.64) & 83.27 (0.43) & 82.96 (0.53) \\\hline
adapter-e2e & 78.54 (0.09) & 81.39 (0.01) & 79.91 (0.05)  \\ 
modular-swinbert & 83.29 (0.23) & 83.58 (0.24) & 83.42 (0.20) \\
modular-kosmos & 82.46 (0.81) & 83.54 (0.28) & 82.98 (0.33) \\ \hline
upper-bound & 83.40 (1.87) & 84.45 (2.35) & 83.91 (1.82) \\ \bottomrule
\end{tabular}
}
\end{center}
\caption{BERTScore for our model as well as all comparison models.} \label{tab:bertscore-comp-models}
\end{table}

\begin{table}
\resizebox{\columnwidth}{!}{
\begin{tabular}{@{}llll@{}}
\toprule
 & bs-precision & bs-recall & bs-f1 \\
\midrule
nocaptions & 83.30 (0.20) & 83.68 (0.22) & 83.48 (0.10) \\
w/o video & 83.62 (0.09) & 83.55 (0.23) & 83.57 (0.13) \\
w/o reorder & 82.82 (0.18) & 83.52 (0.06) & 83.16 (0.10) \\ 
w/o transcript & 83.10 (0.15) & 81.45 (0.09) & 82.25 (0.08) \\\hline
modular-kosmos & 82.46 (0.81) & 83.54 (0.28) & 82.98 (0.33) \\ \hline
\end{tabular}
}

\caption{BERTScore for our our ablation settings.} \label{tab:bertscore-ablation-settings}
\end{table}

\section{Significance Calculations} \label{app:significance-calculations}
The improvements in our model over comparison models, as shown in Table~\ref{tab:main-results} are statistically significant at $\alpha<0.01$. We now calculate an upper bound on all metrics to establish this, for an unpaired unequal variances T-test, i.e.. a Welch test. The t-value for such a test is calculated as 

\begin{equation} \label{eq:welch-t-value}
t_w = \frac{\mu_1 - \mu_2}{\sqrt{\sigma_1^2 + \sigma_2^2}}\,,
\end{equation}

where $\mu_i$ and $\sigma_i^2$ are the mean and variance, respectively, for the $i$th sample. Let $\mu_1, \sigma_1^2$ denote the mean and variance for our model, and $\mu_2, \sigma_2^2$ those for some comparison model. Excluding Llama-7B as a special case, which we consider below, then, for ROUGE1, $\mu_1 0.6$, and $\sigma_2 \leq 0.42$. Therefore, the denominator in \eqref{eq:welch-t-value} is $\leq  \approx 0.328$. Also, $\mu=44.86$ $\mu_2 \leq 42.24$. Thus,
\[
t_w \geq \frac{44.86 - 42.24}{\sqrt{\frac{0.6^2}{5} + \frac{0.42^2}{5}}} \approx 4.47\,.
\]
Similarly, for ROUGE2
\[
t_w \geq \frac{11.83 - 10.32}{\sqrt{\frac{0.15^2}{5} + \frac{0.34^2}{5}}} \approx 4.78\,.
\]
and for ROUGE-Lsum
\[
t_w \geq \frac{42.49 - 40.40}{\sqrt{\frac{0.56^2}{5} + \frac{0.34^2}{5}}} \approx 4.72\,.
\]
Using the Welch-Satterthwaite equation, the degrees of freedom $\nu$ can also be lower-bounded at 5, giving a critical value, for $\alpha=0.1$ of 3.37, which is less than our t-value for each for ROUGE1, ROUGE2 and ROUGE-Lsum.

For FActScore, we do not run multiple random seeds because of the financial cost, but if we compare distributions over individual facts, we see all comparisons are very highly significant. Each setting involves $\sim15,000$ facts, with the `no transcript'' ablation having fewer, $\sim 4,000$. Expressing fact-precision and fact-recall as fractions rather than percentages, and recalling that Bernoulli distribution with mean $p$ has variance $p(1-p)$, we see that the denominator in \eqref{eq:welch-t-value} is upper-bounded by $\frac{2}{4000}$, so the t-value is lower-bounded by $2000(\mu_1 - \mu_2)$. This is in the hundreds for all $\mu_1, \mu_2$ from the factscore-precision and factscore-recall columns in Table~\ref{tab:main-results}, which is far above the critical value of 2.33.

\section{Correlation Between Metrics} \label{app:corr-between-metrics}
\begin{figure}[t]
    \begin{center}
   \includegraphics[width=0.5\textwidth]{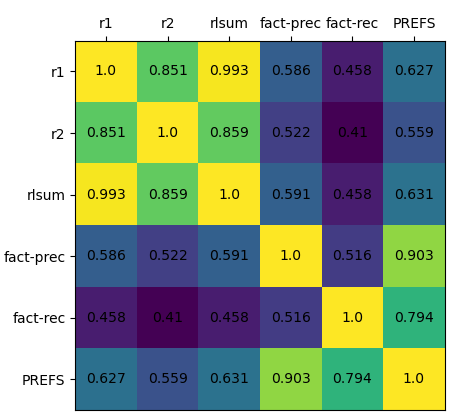}
  \end{center}
    \caption{Correlation between all pairs of metrics that we report in Section \ref{sec:experimental-eval}. All are weakly correlated, with a stronger correlation between the different varieties of ROUGE.}
    \label{fig:metrics-correlation}
\end{figure}

Figure~\ref{fig:metrics-correlation} shows the pairwise correlation (Pearson) between all metrics that we report in Section \ref{sec:experimental-eval}. This is taken across all data points in all settings. All metrics are at least weakly correlated with each other, with the strongest correlations between the different varieties of ROUGE. This suggests that our \textsc{PRisma} metric captures information outside of what is captured by ROUGE.

\onecolumn
\section{Human Evaluation Setup} \label{app:human-eval-guidelines}

We request that each participant read the following guidelines before ranking the summaries.
\begin{quote}
When comparing summaries, you should take into account the following three points, in order of importance. (1) Firstly, most importantly, does it give plenty of detail about the events in the episode, rather than repeating the same information multiple times? (2) Does it report events and characters that really are in the episode, rather than irrelevant information, e.g. about when and where the episode was aired or reviews it received? (3) Is the text well-written, containing full properly-formed sentences devoid of spelling, grammar, and punctuation errors? 

The following is an example of a good summary:

Clint goes to Robert's home and has two guys assault him for what he did to Jessica. But he tells Robert he still gets to keep the job that Clint has enabled him to keep at LU and James can keep his grant. Meanwhile, Jessica and Brody are happily planning their future together with their new baby but agree to get the paternity test so that they know. In the place where Eli has kept her, Dani manages to untie herself from the chair and rushes to the other room in an attempt to save Starr and Hope, assuming they are on the other side. But when she gets through, she is shocked and elated to see her mom. They cry and rejoice but are horrified to find out that Eli, whom they've both seen as a trusted family member and upstanding citizen, is a serial killer and psycho. She admits she does not remember much but is going to get better and be with her daughter and husband. Dani unties her mom and they are ready to go home while Eli is gone. Eli goes to find unconscious Greg in his hospital room and is ready to inject a lethal dose and kill him knowing what Greg was planning, until Shaun catches him and he runs. Greg is still unconscious however, and Eli escapes. And he comes back in time to prevent Tea and Dani from escaping. Meanwhile, Hannah holds Starr and baby Hope captive and is determined to keep Starr away from Cole for as long as it takes for him to forget all about Starr and fall in love with Hannah. But when she returns to Marty's, James notices that she is carrying his bullet on the chain in her purse. He finds that very odd since it was last in Starr's possession. He also finds it very odd when Nate informs him and Robert that Eli called and let him, Todd and Blair all talk to Dani but made no mention of Starr and nobody knows where Starr and Hope are. He senses they are in danger and is determined to save them.

The following is an example of a bad summary, which repeats the same point again and again instead of covering all the information in the episode:

Ford is beaten up by two men that Clint hires, but Clint does not fire him from his job at NU, even though Ford expects that he will be fired. Clint is angry because Ford took advantage of his daughter, Jessica, while she was vulnerable. Jessica and Brody are worried about who the father of their baby is, so they are doing a paternity test. They are doing the paternity test so they can find out who the father is.They are worried about who the father is, but they are going to do a paternity test so they can find out who the father is.They are worried about who the father is, but they are going to do a paternity test so they can find out who the father is.The paternity test will tell them who the father is, and they are worried about it because then they will find out who the father is. They are doing the paternity test to find this out, and find out who the father is with the paternity test. They are very worried about it, but are doing a paternity test to find out for sure. This will tell them who the father is.

The following is an example of a bad summary that is poorly written, discusses irrelevant material and contains garbled sentences:

Clint pays two large men to come with him to Ford's apartment and beat him up for what Ford@; Ford did, the apartment downstairs, Ford did to his daughter. Ford thinks he will be fired from NU but Clint says he is not. Ford: I guess I'm gonna be out of a job, huh? At The  The..  the hospital, Brody meets Jessica to do a paternity test. .. Brody and.. The hospital. The test is ready to go. The episode airs at 9 p.m. ET/PT on Thursday, December 14. The series finale airs on December 15 at 8 p.M. ET on ABC. Jessica: whatever happens, we'll raise this baby together,. Brody: I'm glad you think so honey. Jessica: let's just do the test to be sure. Dani: Mom? I never thought I'd find you here, I thought you were dead. Tea: I'm so happy that you're here. Now let's get out of here before Eli comes back. The season finale airs at 10 p. M.E. on December 16 at 8:30 p.S. ET.M ET on E.C. ET at the Palace Hotel in New York City, New York, and on December 18 at 9 a.C.. The finale will air at 9 P.M.. The season. The finale airs. The premie.. The premieThe finale he series.. The  season finale. The.. season finale..The. season finale aired.. The. finale airs.. September 14 at 9:00 p. m. ET, September 14, 2013,
\end{quote}
The `good summary` example is a gold summary from the train set. The `bad summary` examples are composed by us.

We also ask each participant a question about the contents the episode they are assigned, to check that they have watched and understood the full episode. The list of episodes, and corresponding questions and answers are shown in Table \ref{tab:human-eval-attn-check-qs}.

\begin{table*}[]
    \centering
    \begin{tabular}{c|c|c}
         Episode &  Question & Answer \\
         \thead{Port Charles, 05-06-03} & \thead{what sketch does Kevin show to Elizabeth?} & \thead{a sketch of her.} \\
         \thead{The Bold and the Beautiful, 05-22-06} & \thead{what does Stephanie promise to never do again \\ if Brooke comes back to the company?} & \thead{disrespect her} \\
         \thead{Guiding Light, 02-28-05} & \thead{what flavour ice cream did the grandson, \\ Zach, get all over his face?} & \thead{chocolate fudge swirl} \\
         \thead{One Life to Live, 01-06-12} & \thead{What are the names of Allison's two daughters\\  that Vicki faked the paternity test for?} & \thead{Natalie and Jessica} \\
         \thead{The Bold and the Beautiful, 04-27-06} & \thead{what is Stephanie trying to \\ convince Bridget not to do?} & \thead{take her grandson to Italy} \\
         \thead{Guiding Light, 09-06-06} & \thead{who is Tammy in love with?} & \thead{Johnathan} \\
         \thead{As the World Turn, 01-06-05} & \thead{who does Henry say he's in love with?} & \thead{Katie} \\
         \thead{Guiding Light, 01-24-05} & \thead{who does Harley think Beth might have done?} & \thead{Killed Phillip} \\
         \thead{The Bold and the Beautiful, 01-02-15} & \thead{who is Rick convincing to sign the papers?} & \thead{Eric, his dad}
    \end{tabular}
    \caption{The questions, and corresponding answers, for each episode that we ask particpants in order to check they fully watched and understood the video. Episode names are in the format <show-name> <date-aired (MM-DD-YYYY)>.}
    \label{tab:human-eval-attn-check-qs}
\end{table*}
\end{document}